\documentclass[conference]{IEEEtran}
\IEEEoverridecommandlockouts
\usepackage{cite}
\usepackage{amsmath,amssymb,amsfonts}
\usepackage{algorithmic}
\usepackage[dvipdfmx]{graphicx}
\usepackage{textcomp}
\usepackage{xcolor}  
\usepackage{booktabs}
\usepackage{makecell}
\usepackage{here}
\usepackage{url} 
\usepackage{multicol}
\usepackage{adjustbox} 
\usepackage{multirow}

\usepackage[linesnumbered,ruled,vlined]{algorithm2e}

\makeatletter
\newcommand{\newlineauthors}{%
  \end{@IEEEauthorhalign}\hfill\mbox{}\par
  \mbox{}\hfill\begin{@IEEEauthorhalign}
}
\makeatother

\def\BibTeX{{\rm B\kern-.05em{\sc i\kern-.025em b}\kern-.08em
    T\kern-.1667em\lower.7ex\hbox{E}\kern-.125emX}}

\usepackage{tikz} 
\usetikzlibrary{shapes.geometric, arrows, positioning, calc} 
\tikzstyle{data} = [
    rectangle, rounded corners,
    minimum width=2cm, minimum height=1cm,
    text centered, align=center,
    draw=black, fill=gray!20
]

\tikzstyle{signal_encoder} = [
    rectangle,
    minimum width=1.6cm, minimum height=1cm,
    text centered, align=center,
    draw=black, fill=blue!15
]
\tikzstyle{text_encoder} = [
    rectangle,
    minimum width=2.cm, minimum height=1cm,
    text centered, align=center,
    draw=black, fill=blue!15
]

\tikzstyle{attention} = [
    rectangle,
    minimum width=1.5cm, minimum height=1cm,
    text centered, align=center,
    draw=black, fill=green!20
]

\tikzstyle{process} = [
    rectangle,
    minimum width=2cm, minimum height=1cm,
    text centered, align=center,
    draw=black, fill=orange!20
]

\tikzstyle{arrow} = [
    thick, ->, >=stealth, draw=black
]

\begin{document}

\title{LaSTR: Language-Driven Time-Series Segment Retrieval}

\author{\IEEEauthorblockN{Kota Dohi, Harsh Purohit, Tomoya Nishida,
Takashi Endo, Yusuke Ohtsubo, Koichiro Yawata,\\
Koki Takeshita, Tatsuya Sasaki, Yohei Kawaguchi}
\IEEEauthorblockA{\textit{Research and Development Group, Hitachi, Ltd.}}
}

\maketitle

\begin{abstract}

Effectively searching time-series data is essential for system analysis, but existing methods often require expert-designed similarity criteria or rely on global, series-level descriptions.
We study language-driven segment retrieval: given a natural language query, the goal is to retrieve relevant local segments from large time-series repositories.
We build large-scale segment--caption training data by applying TV2-based segmentation to LOTSA windows and generating segment descriptions with GPT-5.2, and then train a Conformer-based contrastive retriever in a shared text--time-series embedding space.
On a held-out test split, we evaluate single-positive retrieval together with caption-side consistency (SBERT and VLM-as-a-judge) under multiple candidate pool sizes.
Across all settings, LaSTR outperforms random and CLIP baselines, yielding improved ranking quality and stronger semantic agreement between retrieved segments and query intent.
\end{abstract}


\section{Introduction}
\label{sec:intro}
With the rapid growth of the Internet of Things (IoT), the volume of sensor time-series data to be analyzed has increased dramatically. In many industrial settings, retrieving time-series data that exhibit user-desired behaviors from massive repositories still requires substantial manual effort by domain experts~\cite{Imani2019}. To reduce this dependence on specialized personnel and enable less experienced analysts to conduct exploratory analysis, prior studies have investigated natural-language-based retrieval of sensor time series~\cite{Imani2019, ito2024clasp,dohi2025retrieval}.

However, most existing approaches perform retrieval based on the global shape of an entire time series. In practice, analysts often seek segments where a query-relevant local pattern is salient relative to the rest of the series (e.g., a prominent spike followed by a transient downward drift), rather than coarse, series-level descriptions such as “an overall increasing trend” over the full duration~\cite{Yeh2016MatrixProfileI}. Global-shape retrieval does not support this segment-centric notion of salience, limiting the applicability of existing methods to real-world workflows.

In this work, we propose LaSTR, a framework for language-driven time-series segment retrieval. Given a natural-language query, the system retrieves the most relevant segments from a large collection of time-series data. We first generate candidate segments by applying piecewise linear approximation to time series from diverse domains~\cite{Keogh2001OnlineSegmentation}. For each segment, we then leverage a vision–language model (VLM) to generate a natural-language description that highlights the segment’s distinctive temporal patterns and characterizes the segment relative to its surrounding context in the full series~\cite{OpenAI2025GPT52SystemCard}. Using the resulting (segment, description) pairs, we train a contrastive learning model that projects time-series segments and natural-language queries into a shared embedding space~\cite{Chen2020SimCLR}. This enables efficient retrieval of time-series segments that match the user’s intent expressed in natural language.

\section{Relation to prior work}
Retrieval of time-series patterns from large databases has been extensively studied in data mining and signal processing~\cite{Christos1994, Keogh2001Dimensionality,keogh2003need}.
In these approaches, retrieval quality strongly depends on the definition of an appropriate similarity measure, including shape-based metrics tailored to specific domains or use cases \cite{nakamura2013shape}.
This often makes effective retrieval reliant on expert knowledge and careful metric design.

To reduce this dependency, recent work leverages natural language supervision to connect time-series data with textual concepts, enabling natural-language-based access to time-series data~\cite{Imani2019,ito2024clasp}.
Related efforts also explore linking language models and sensor signals in adjacent settings such as zero-shot recognition~\cite{zhou2023tentconnectlanguagemodels}.
However, most language-based time-series retrieval setups emphasize coarse or global representations by treating an entire series as a single instance, while practical analysis often requires retrieving short, temporally localized events within long recordings~\cite{Yeh2016MatrixProfileI}.

In contrast, we study language-driven time-series segment retrieval, where the goal is to retrieve relevant segments from large-scale repositories given a free-form natural language query.
Our approach leverages automatically produced segment-level descriptions as pseudo-labels to train a contrastive model that embeds time-series segments and text queries into a shared space, enabling efficient retrieval of localized temporal behaviors at scale.

\section{Problem Statement}
\label{sec:problem_statement}

We study language-driven time-series segment retrieval.
A large set of candidate segments is assumed to be available (e.g., provided or generated by a preprocessing step),
and the goal is to retrieve relevant segments from this candidate pool using a natural-language query.

Let $\mathcal{X}=\{x^{(n)}\}_{n=1}^{N}$ be a collection of univariate time series, where
$x^{(n)}=(x_t^{(n)})_{t=1}^{T_n}$ and $x_t^{(n)}\in\mathbb{R}$.
For each series $x^{(n)}$, candidate segment boundaries are given as a set of discrete index pairs
$\mathcal{I}^{(n)}=\{(a_m^{(n)}, b_m^{(n)})\}_{m=1}^{M_n}$ with
$1 \le a_m^{(n)} < b_m^{(n)} \le T_n$,
where $a_m^{(n)}$ and $b_m^{(n)}$ are integer time indices.
A candidate segment specification is defined by the parent-series index and a boundary pair.
The global set of candidate segment specifications is
\begin{equation}
\mathcal{S}
=\{(n,a_m^{(n)},b_m^{(n)}) \mid n=1,\dots,N,\; m=1,\dots,M_n\}.
\end{equation}
For a specification $s=(n,a,b)\in\mathcal{S}$, the corresponding segment (subsequence) is
$x^{(n)}[a:b] = (x_t^{(n)})_{t=a}^{b}$.

Given a natural-language query $q$ specifying the desired temporal behavior and its relative salience within the parent series,
the goal is to retrieve the $K$ most relevant specifications from $\mathcal{S}$.
To this end, a scoring function $r\!\left(x^{(n)}, a, b \mid q\right)$ is learned to assign a relevance score to each candidate specification $(n,a,b)\in\mathcal{S}$ conditioned on $q$.
At test time, the retrieved set $\hat{\mathcal{S}}_{K}(q)$ is obtained by selecting the top-$K$ scoring specifications.
Let $\operatorname*{arg\,topK}$ denote the operator that returns the set of $K$ specifications with the highest scores:
\begin{equation}
\label{eq:retrieval}
\hat{\mathcal{S}}_{K}(q)
= \operatorname*{arg\,topK}_{(n,a,b) \in \mathcal{S}} \; r\!\left(x^{(n)}, a, b \mid q\right).
\end{equation}

\section{Method}
\subsection{Large-Scale Segment--Caption Pair Generation}
\label{ssec:segment-caption-generation}

\subsubsection{Data and preprocessing.}
We extract fixed-length windows of length $L_w$ from longer raw time series and treat each window as a sample $x^{(n)}$.
Each window is min--max normalized to $[0,1]$, and we denote the normalized window by $\bar{x}\in[0,1]^{L_w}$.

\subsubsection{Segment generation.}
To obtain candidate segments, we fit a piecewise-linear approximation to each normalized window $\bar{x}$ and use sharp curvature changes as boundary cues.
Specifically, we estimate a denoised trend $\hat{x}$ by second-order total-variation (TV2) regularization:
\begin{equation}
\hat{x}=\arg\min_{u\in\mathbb{R}^{L_w}}
\left(\|u-\bar{x}\|_2^2+\lambda\|D^2 u\|_1\right),
\label{eq:tv2}
\end{equation}
where $(D^2 u)_t=u_{t+1}-2u_t+u_{t-1}$ is the discrete second difference.
The resulting $\hat{x}$ is piecewise linear; we therefore detect boundary candidates from the second difference
$\Delta^2\hat{x}_t=\hat{x}_{t+1}-2\hat{x}_t+\hat{x}_{t-1}$.
We set a threshold $\theta$ based on the standard deviation of $\Delta^2\hat{x}$:
\begin{equation}
\theta \triangleq 3\,\sigma_{\Delta^2\hat{x}},
\qquad
\sigma_{\Delta^2\hat{x}}
=\sqrt{\frac{1}{L_w-2}\sum_{t=2}^{L_w-1}
\left(\Delta^2\hat{x}_t-\overline{\Delta^2\hat{x}}\right)^2}.
\end{equation}
Indices satisfying $\left|\Delta^2\hat{x}_t\right|>\theta$ are treated as change points. The detected points are sorted in time and converted into contiguous boundary pairs $(a_m, b_m)$.
For each window, we set $\lambda$ adaptively so that the number of detected segments does not exceed $M_{\max}$.

\subsubsection{Caption generation with VLM.}
\begin{figure}[t]
  \centering
  \includegraphics[width=\linewidth]{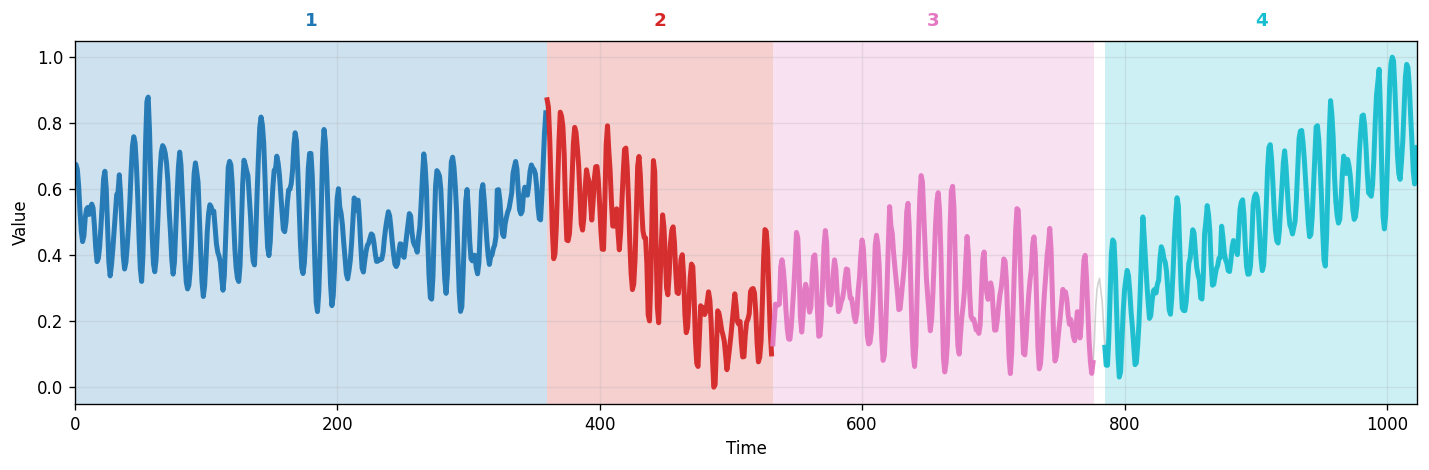}
  \vspace{0.4em}
  \begin{minipage}{0.98\linewidth}
  \footnotesize
  \raggedright
  \textbf{(1)} Choppy oscillations drift slightly upward, continuing a broadly sideways range.\\
  \textbf{(2)} Volatile slide with sharp dips, marking a breakdown from prior stability.\\
  \textbf{(3)} Noisy sideways consolidation with small rebounds, pausing after the selloff.\\
  \textbf{(4)} Steady climb with strong swings, launching a sustained recovery uptrend.
  \end{minipage}
  \caption{Example VLM input and corresponding outputs for segment captioning. The four captions are generated in the same order as the segment indices shown in the plot.}
  \label{fig:example_vlm}
\end{figure}

Only segments longer than a minimum-length threshold $L_{\min}$ are sent to the VLM for captioning.
For each window, we render a single-panel plot as the VLM input: the full window is drawn in light gray, and each target segment is overlaid with a semi-transparent colored band and a thicker colored line; the segment order is indicated by a number placed above the segment center (Fig.~\ref{fig:example_vlm}).
We use GPT-5.2 via the Chat Completions API (model identifier: \texttt{gpt-5.2}; no ``pro'' or ``thinking'' suffix).
The prompt requests a JSON array of segment captions in index order.
Detailed reproducibility information (full prompt templates, API parameters, retry/timeout policy, and response validation rules) is provided on the project page~\cite{dohi2026lastrproject}.
Each caption $c_m^{(n)}$ is stored together with the corresponding candidate segment specification $(n,a_m^{(n)}, b_m^{(n)})\in\mathcal{S}$.
This yields a large collection of segment--caption pairs for contrastive training.

\subsection{Segment-Level Contrastive Learning with Conformer}
We train a segment--caption retrieval model using the segment--caption pairs generated by the procedure described in Section~\ref{ssec:segment-caption-generation}.
A mini-batch contains $B$ paired examples $\{(\bar{x}_i, a_i, b_i, c_i)\}_{i=1}^{B}$, where $\bar{x}_i\in[0,1]^{L_w}$ is a normalized window, $(a_i,b_i)$ are boundary indices within $\bar{x}_i$, and $c_i$ is the corresponding caption.
Given $\bar{x}_i$, a Conformer-based time-series encoder $f_\theta$~\cite{gulati20_interspeech} produces frame-level representations
\begin{equation}
H_i=f_\theta(\bar{x}_i)\in\mathbb{R}^{L_w\times d}.
\end{equation}
We design the encoder to produce frame-level representations that integrate both local dynamics and broader context.
Concretely, the encoder preserves temporal resolution so that $(H_i)_t$ remains aligned with input frame $t$ while incorporating contextual information via self-attention.
This alignment enables segment-level pooling based on boundary indices $(a_i,b_i)$, yielding a segment embedding that reflects not only the local behavior within $[a_i,b_i]$ but also its surrounding context in the window.
The segment feature is obtained by average pooling over the boundary interval:
\begin{equation}
h_i = \frac{1}{b_i - a_i + 1} \sum_{t=a_i}^{b_i} (H_i)_t.
\end{equation}

Let $e_{\mathrm{text}}$ denote a text encoder, and let $g_{\mathrm{seg}}$ and $g_{\mathrm{text}}$ denote projection heads for segment and text features, respectively.
We project both modalities into a shared space and apply $\ell_2$ normalization:
\begin{equation}
z_i=\frac{g_{\mathrm{seg}}(h_i)}{\|g_{\mathrm{seg}}(h_i)\|_2},
\qquad
u_i=\frac{g_{\mathrm{text}}(e_{\mathrm{text}}(c_i))}{\|g_{\mathrm{text}}(e_{\mathrm{text}}(c_i))\|_2}.
\end{equation}
The temperature-scaled similarity matrix is
\begin{equation}
\Psi_{ij}=\frac{z_i^\top u_j}{\tau},
\end{equation}
where $\tau>0$ is a temperature parameter.
We optimize a symmetric InfoNCE loss~\cite{Oord2018CPC}, where the positive pair is the matched index in the mini-batch:
\begin{equation}
\mathcal{L}
= \frac{1}{2}\Bigl(\mathrm{CE}(\Psi, y) + \mathrm{CE}(\Psi^\top, y)\Bigr),
\qquad y_i=i,
\end{equation}
where $\mathrm{CE}$ is cross-entropy over rows and $y$ selects the diagonal match for each pair.

\subsection{Retrieval at Test Time}
At inference, a query caption $q$ is encoded into the shared space:
\begin{equation}
u_q=\frac{g_{\mathrm{text}}(e_{\mathrm{text}}(q))}{\|g_{\mathrm{text}}(e_{\mathrm{text}}(q))\|_2}.
\end{equation}
For each candidate segment specification $s=(n,a,b)\in\mathcal{S}$, we compute its normalized embedding by applying the same encoder and pooling procedure as in training:
\begin{equation}
z(s)=\frac{g_{\mathrm{seg}}\!\left(\frac{1}{b-a+1}\sum_{t=a}^{b} f_\theta(\bar{x}^{(n)})_t\right)}{\left\|g_{\mathrm{seg}}\!\left(\frac{1}{b-a+1}\sum_{t=a}^{b} f_\theta(\bar{x}^{(n)})_t\right)\right\|_2}.
\end{equation}
The relevance score is the cosine similarity in the shared space,
$r(s\mid q)=u_q^\top z(s)$,
and the retrieval output is the top-$K$ scored specifications:
\begin{equation}
\hat{\mathcal{S}}_K(q)=\operatorname*{arg\,topK}_{s\in\mathcal{S}} r(s\mid q).
\end{equation}

\section{Experiments}
\subsection{Dataset}
We construct the dataset from LOTSA~\cite{woo2024unified}, a large-scale time-series collection spanning diverse real-world domains, including energy, transportation, weather/climate, finance, healthcare, and web/IT services.
In this study, we use 170 subsets from LOTSA and generate segment--caption pairs using the pipeline in Section~\ref{ssec:segment-caption-generation}.
The 170 subsets are randomly partitioned into 155/5/10 subsets for the train/validation/test splits, with no overlap of subsets across splits.

Let $N_{\text{subset}}$ denote the number of time series in a subset.
For each subset, we sample $N_{\text{target}}=1000$ windows of fixed length $L_w=1024$.
To mitigate over-representation by a small number of long series, windows are allocated approximately uniformly across the $N_{\text{subset}}$ series:
each series is assigned $\lfloor N_{\text{target}}/N_{\text{subset}}\rfloor$ windows, and one additional window is assigned to a number of series so that the total equals $N_{\text{target}}$.

Windows are extracted using a sliding window of length $1024$ with an initial stride of $1024$; the stride is reduced only when necessary to satisfy the per-series allocation.
If a series cannot provide a sufficient number of distinct windows, duplicate windows are permitted.
Series shorter than $1024$ points are linearly interpolated to length $1024$ prior to sampling.

For each sampled window, candidate segments are generated using the procedure in Section~\ref{ssec:segment-caption-generation}, and captions are produced with GPT-5.2 as the VLM~\cite{OpenAI2025GPT52SystemCard}.
For TV2 segmentation (Eq.~\ref{eq:tv2}), the regularization parameter is initialized at $\lambda=100$ and multiplied by 10 until the number of detected segments is at most $M_{\max}=6$.
Only segments of length at least $L_{\min}=50$ are sent to the VLM for captioning.

With 155/5/10 subset splits and 1{,}000 windows per subset, the target window counts are 155{,}000/5{,}000/10{,}000 for train/validation/test, respectively.
This results in 560{,}223/17{,}591/36{,}852 segment--caption pairs.

\subsection{Experimental conditions}
The time-series encoder is a Conformer with four layers, four attention heads, hidden size 128, convolution kernel size 31, dropout 0.2, and half-step residual connections. The text encoder is BERT-base-uncased~\cite{devlin-etal-2019-bert} and is kept frozen throughout training. The outputs of both encoders are projected to a shared 128-dimensional embedding space using two-layer MLP projection heads with ReLU activations. We train the model with a contrastive objective using temperature $\tau=0.07$. We optimize using AdamW~\cite{loshchilov2019decoupled} with learning rate $1\times10^{-4}$ and weight decay 0.05, using 500 warmup steps. Gradients are clipped to a maximum norm of 1.0. We use a batch size of 512 and train for 100 epochs. The best checkpoint is selected by the highest validation Acc@1, computed within each batch on the validation split: a segment is counted as correct if its paired caption is the top-ranked caption among all captions in the same batch.

\subsection{Evaluation}
We evaluate each method on the test split.
All methods are evaluated using the same fixed query set and the same query-specific candidate pools described below.

\textbf{Query set.}
We construct a query set by uniformly sampling $N_q=100$ segment--caption pairs from the test split.
Each query $q_i$ is a single caption and is paired with exactly one ground-truth segment specification $s_i$ (single-positive).

\textbf{Query-specific candidate pool.}
For each query $q_i$ with ground-truth segment specification $s_i=(n_i,a_i,b_i)$, a candidate pool is constructed at the time-series window level.
The parent window $x^{(n_i)}$ indicated by $s_i$ is included in the pool, and additional windows are sampled uniformly at random from the remaining test windows until the pool contains $N_{\mathrm{pool}}$ windows in total.
All candidate segment specifications from the pooled windows constitute the candidate set $\mathcal{C}_i$.
Multiple pool sizes are evaluated ($N_{\mathrm{pool}}\in\{100,1000,10000\}$).

\begin{table}[t]
\centering
\caption{Retrieval performance evaluated in single-positive segment retrieval setting with different pool sizes.}
\begin{tabular}{llcccc}
\toprule
Pool & Method & Recall@1 & Recall@5 & Recall@10 & mAP \\
\midrule
100   & Random & 0.000 & 0.000 & 0.000 & 0.009 \\
100   & CLIP   & 0.010 & 0.020 & 0.030 & 0.027 \\
100   & LaSTR   & \textbf{0.240} & \textbf{0.710} & \textbf{0.840} & \textbf{0.446} \\
\midrule
1000  & Random & 0.000 & 0.000 & 0.000 & 0.001 \\
1000  & CLIP   & 0.000 & 0.000 & 0.010 & 0.003 \\
1000  & LaSTR   & \textbf{0.050} & \textbf{0.220} & \textbf{0.310} & \textbf{0.145} \\
\midrule
10000 & Random & 0.000 & 0.000 & 0.000 & 0.000 \\
10000 & CLIP   & 0.000 & 0.000 & 0.000 & 0.001 \\
10000 & LaSTR   & \textbf{0.030} & \textbf{0.070} & \textbf{0.130} & \textbf{0.065} \\
\bottomrule
\end{tabular}
\label{tab:pool_sweep_retrieval}
\end{table}

\textbf{Ranking.}
We compute cosine similarity between the query embedding $u_i$ and each candidate time-series segment embedding $z_j$ for $j\in\mathcal{C}_i$:
\begin{equation}
\rho_{ij}=u_i^\top z_j.
\end{equation}
Candidates in $\mathcal{C}_i$ are ranked by $\rho_{ij}$.
Let $j_i^\star$ denote the index in $\mathcal{C}_i$ corresponding to the ground-truth segment specification $s_i$, and let $r_i$ denote its rank:
\begin{equation}
r_i = 1 + \left|\left\{j \in \mathcal{C}_i \mid \rho_{ij} > \rho_{i j_i^\star}\right\}\right|.
\end{equation}

\textbf{Metrics.}
\emph{(1) Single-positive segment retrieval.}
We report
\begin{equation}
\mathrm{Recall@}K = \frac{1}{N_q}\sum_{i=1}^{N_q}\mathbb{I}[r_i \le K],
\end{equation}
and
\begin{equation}
\mathrm{mAP} = \frac{1}{N_q}\sum_{i=1}^{N_q}\frac{1}{r_i},
\end{equation}

\emph{(2) Sentence-BERT (SBERT)-based caption consistency (Top-$K$)~\cite{reimers2019sentence}.}
Let $c_{i,(r)}$ denote the caption paired with the rank-$r$ retrieved segment for query $i$.
Let $f_{\mathrm{S}}(\cdot)$ be the Sentence-BERT encoder and define the normalized embedding
$e_{\mathrm{S}}(x)=\frac{f_{\mathrm{S}}(x)}{\left\|f_{\mathrm{S}}(x)\right\|_2}$.
We report
\begin{equation}
\mathrm{Mean\ SBERT@}K=\frac{1}{N_q}\sum_{i=1}^{N_q}
\left(\frac{1}{K}\sum_{r=1}^{K}
e_{\mathrm{S}}(q_i)^\top e_{\mathrm{S}}\!\left(c_{i,(r)}\right)\right).
\end{equation}

\emph{(3) VLM-as-a-judge (Top-$K$).}
For each query, the top-$K$ retrieved segments are visualized and evaluated by a VLM (GPT-5.2).
Two judge prompts are used.
The first prompt requests a 5-level relevance rating, yielding scores $v_{i,r}\in\{1,2,3,4,5\}$ for rank $r$ of query $q_i$.
The second prompt requests a binary decision (match / non-match), yielding labels $\ell_{i,r}\in\{0,1\}$.
The full judge prompts and inference settings are provided on the project page~\cite{dohi2026lastrproject}.
We report
\begin{equation}
\mathrm{VLM\ Score@}K=\frac{1}{N_q}\sum_{i=1}^{N_q}\left(\frac{1}{K}\sum_{r=1}^{K}v_{i,r}\right),
\end{equation}
and
\begin{equation}
\mathrm{VLM\ Precision@}K=\frac{1}{N_q}\sum_{i=1}^{N_q}\left(\frac{1}{K}\sum_{r=1}^{K}\ell_{i,r}\right).
\end{equation}

\textbf{Baselines and compared methods.}
All methods are evaluated under the same query set and query-specific candidate pools.
The random baseline ranks candidates by a random permutation of $\mathcal{C}_i$.
Another baseline uses CLIP~\cite{Radford2021CLIP}\footnote{Model ID: \texttt{openai/clip-vit-base-patch32}.}; each candidate segment is rendered as a single-panel plot with segment highlighting, encoded with CLIP image embeddings, and ranked against CLIP text embeddings of query captions (templated as ``The highlighted segment can be described as: \{caption\}'').
The LaSTR (Conformer-based) model ranks candidates using the learned segment and text encoders in the shared embedding space.

\begin{table}[t]
\centering
\caption{SBERT and VLM-as-a-judge scores (Top-10) across candidate pool sizes.}
\begin{tabular}{llccc}
\toprule
Pool & Method & \makecell{Mean SBERT\\@10} & \makecell{VLM Score\\(1--5)@10} & \makecell{VLM Precision\\(binary)@10} \\
\midrule
100   & Random & 0.496 & 3.550 & 0.351 \\
100   & CLIP   & 0.505 & 3.631 & 0.433 \\
100   & LaSTR   & \textbf{0.625} & \textbf{4.037} & \textbf{0.719} \\
\midrule
1000  & Random & 0.489 & 3.521 & 0.327 \\
1000  & CLIP   & 0.505 & 3.690 & 0.437 \\
1000  & LaSTR   & \textbf{0.639} & \textbf{4.122} & \textbf{0.804} \\
\midrule
10000 & Random & 0.498 & 3.511 & 0.328 \\
10000 & CLIP   & 0.507 & 3.732 & 0.433 \\
10000 & LaSTR   & \textbf{0.634} & \textbf{4.110} & \textbf{0.839} \\
\bottomrule
\end{tabular}
\label{tab:pool_sweep_caption}
\end{table}

\begin{figure}[t]
\centering
\includegraphics[width=\linewidth]{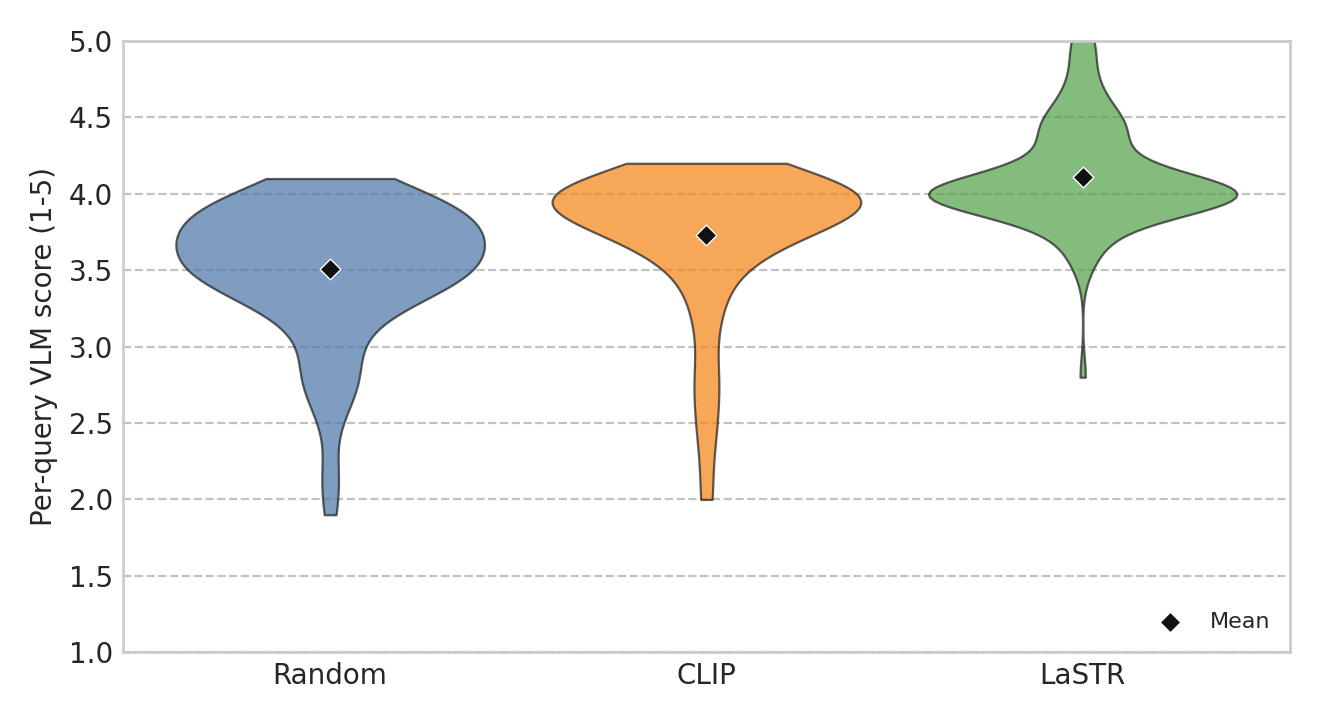}
\caption{Distribution of per-query mean VLM scores over top-10 retrieved segments (test split, pool size $=10000$, 5-point scale).}
\label{fig:vlm_score_distribution}
\end{figure}

\begin{figure*}[t]
\centering
\footnotesize
\setlength{\tabcolsep}{6pt}
\begin{tabular}{m{0.47\textwidth}m{0.47\textwidth}}
\begin{minipage}[t]{\linewidth}
\raggedright\textbf{Query:} Bumpy recovery from lows, rebounding within the prior broader decline.
\begin{center}
\includegraphics[width=0.88\linewidth]{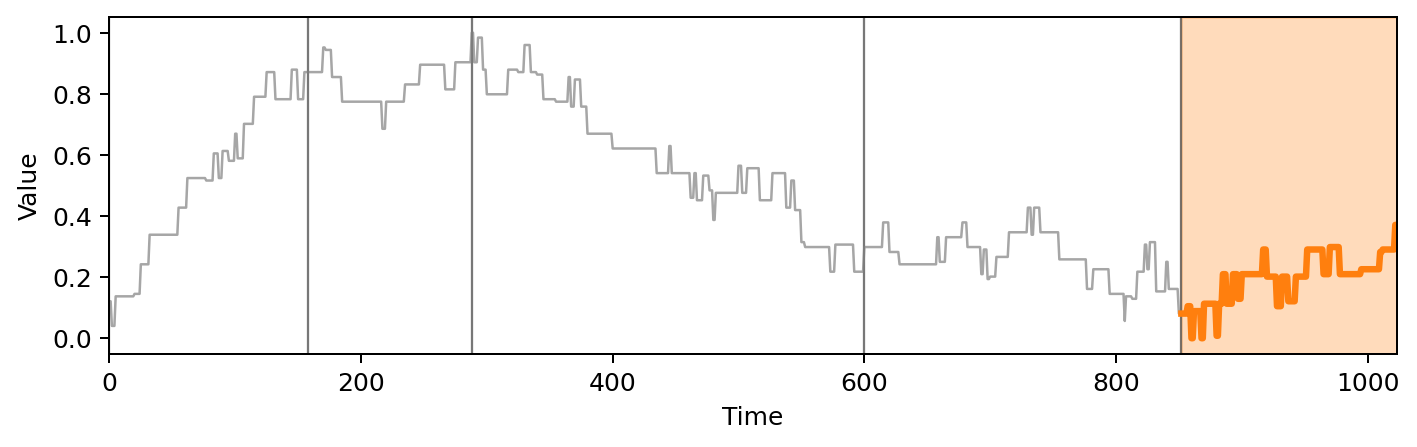}
\end{center}
\end{minipage}
&
\begin{minipage}[t]{\linewidth}
\raggedright\textbf{Query:} Volatile surge then sharp plunge, followed by strong rebound in uptrend.
\begin{center}
\includegraphics[width=0.88\linewidth]{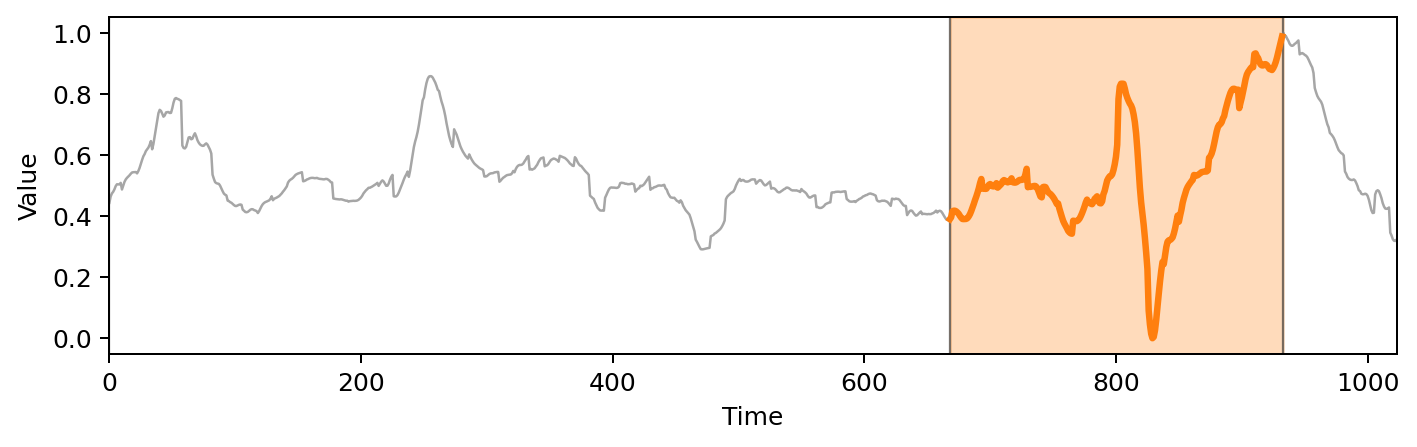}
\end{center}
\end{minipage}
\\[0.6ex]
\begin{minipage}[t]{\linewidth}
\raggedright\textbf{Query:} Steady, noisy decline from peak, marking reversal into sustained downtrend.
\begin{center}
\includegraphics[width=0.88\linewidth]{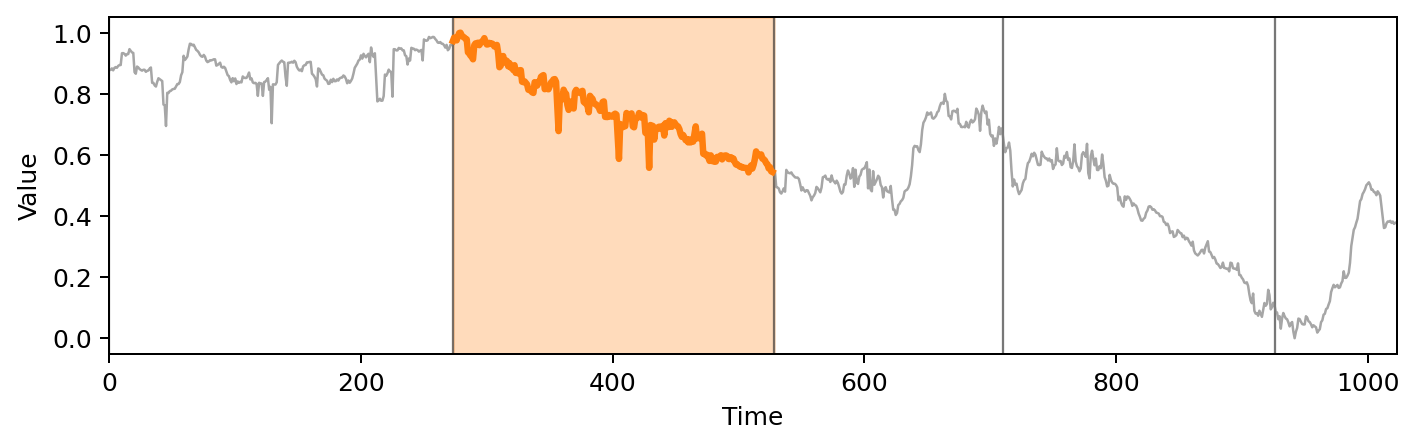}
\end{center}
\end{minipage}
&
\begin{minipage}[t]{\linewidth}
\raggedright\textbf{Query:} Mostly flat near baseline, slight dip, preceding broader upward surge.
\begin{center}
\includegraphics[width=0.88\linewidth]{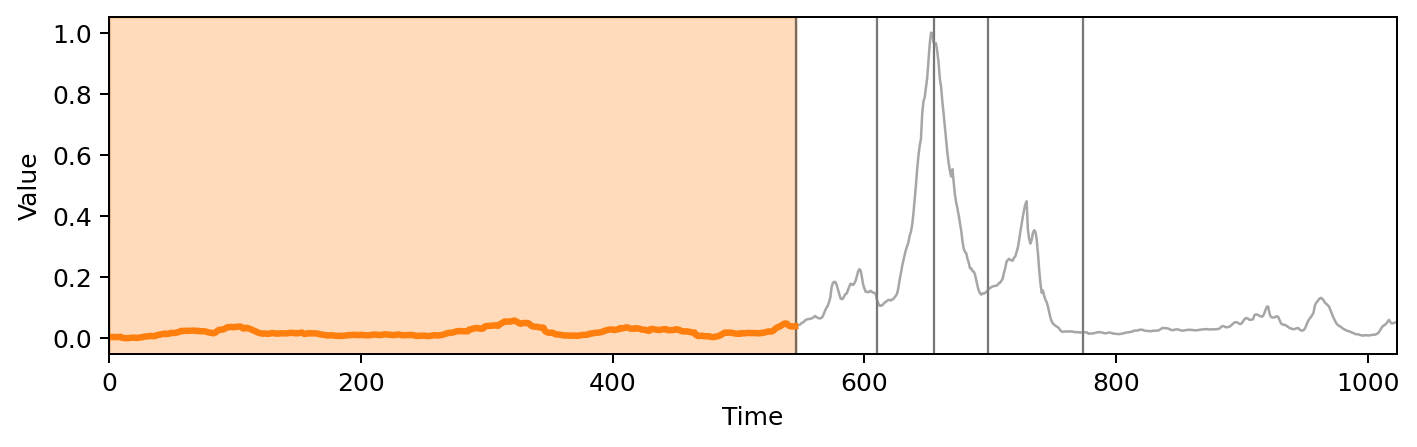}
\end{center}
\end{minipage}
\end{tabular}
\caption{Qualitative retrieval examples on the test split. For each query, the figure shows the query caption and the corresponding rank-1 retrieved time-series window at pool size 10{,}000. Segment boundaries are overlaid, and the retrieved segment is highlighted to indicate the matched region.}
\label{fig:qualitative_rank1_rows}
\end{figure*}

\subsection{Results}
Table~\ref{tab:pool_sweep_retrieval} summarizes single-positive segment retrieval performance in terms of Recall@K and mAP.
Performance decreases as the candidate pool size increases, reflecting the increased difficulty of the retrieval setting.
Across all pool sizes and metrics, LaSTR outperforms both baselines by a substantial margin, demonstrating improved retrieval performance on test subsets that are disjoint from the training data.
The CLIP baseline also surpasses random retrieval, indicating that CLIP captures semantic information that is partially aligned with the query captions.

Table~\ref{tab:pool_sweep_caption} reports caption-side evaluation using SBERT and VLM-as-a-judge.
LaSTR achieves the highest scores for all pool sizes and all three metrics.
In contrast to the single-positive metrics, caption-side scores do not necessarily decrease as the pool size increases; for instance, VLM Precision (binary)@10 is higher at pool size 10{,}000 than at pool size 100.
This observation is consistent with the evaluation setting:
although each query has a single labeled ground-truth segment, multiple additional segments can still be semantically valid for the same query.
As the pool size increases, the candidate set is more likely to contain such valid alternatives, which can then enter the top-$K$ list and sustain or even improve caption-side scores.
The CLIP baseline remains above random under these caption-side metrics.

Figure~\ref{fig:vlm_score_distribution} shows the distribution of per-query mean VLM Score@10 (pool size 10{,}000), which shifts upward from Random to CLIP to LaSTR.
Figure~\ref{fig:qualitative_rank1_rows} shows rank-1 examples where retrieved segments capture both local shape and their relationship to the surrounding context.

\section{Conclusion}
We present LaSTR, a language-driven time-series segment retrieval framework that leverages automatically generated segment--caption pairs and a contrastive learning architecture.
By coupling TV2-based segment extraction with VLM-generated descriptions that capture both segment-local dynamics and their salience relative to the surrounding context, LaSTR supports retrieval of temporally confined behaviors in a context-aware manner.
Experiments on LOTSA-derived test data show consistent improvements over random and CLIP baselines across multiple candidate pool sizes, under both single-positive retrieval metrics and caption-side semantic consistency evaluations.


\bibliographystyle{IEEEtran}
\bibliography{refs}
\end{document}